# Reasoning With Conditional Ceteris Paribus Preference Statements

**Craig Boutilier**
Dept. of Computer Science
University of British Columbia
Vancouver, BC V6T 1Z4
*cebly@cs.ubc.ca*

**Ronen I. Brafman**
Department of Math and CS
Ben-Gurion University
Beer Sheva, Israel 84105
*brafman@cs.bgu.ac.il*

**Holger H. Hoos and David Poole**
Dept. of Computer Science
University of British Columbia
Vancouver, BC V6T 1Z4
*{hoos,poole}@cs.ubc.ca*

## Abstract

In many domains it is desirable to assess the preferences of users in a qualitative rather than quantitative way. Such representations of qualitative preference orderings form an important component of automated decision tools. We propose a graphical representation of preferences that reflects conditional dependence and independence of preference statements under a *ceteris paribus* (all else being equal) interpretation. Such a representation is often compact and arguably natural. We describe several search algorithms for dominance testing based on this representation; these algorithms are quite effective, especially in specific network topologies, such as chain- and tree-structured networks, as well as polytrees.

## 1 Introduction

Preference elicitation is an important aspect of automated decision making. In many application domains, the space of possible actions or decisions available to someone is fixed, with well-understood dynamics; the only variable component in the decision making process are the preferences of the user on whose behalf a decision is being made. This is often the case in domains such as product configuration or medical diagnosis (to name but two).

Extracting preference information from users is generally arduous, and human decision analysts have developed sophisticated techniques to help elicit this information from decision makers [11]. A key goal in the study of computer-based decision support is the construction of tools that allow the preference elicitation process to be automated, either partially or fully. In particular, methods for extracting, representing and reasoning about the preferences of naive users is especially important in AI applications, where users cannot be expected to have the patience (or sometimes the ability) to provide detailed preference relations or utility functions. In applications ranging from collaborative filtering [14] and recommender systems [15] to product configuration [6] to medical decision making [4], typical users may not be able to provide much more than qualitative rankings of fairly circumscribed outcomes.

Ideally, a preference representation for such applications would capture statements that are natural for users to assess, are reasonably compact, and support effective inference (particularly when deciding whether one outcome is preferred to, or dominates, another). In this paper, we explore a network representation of conditional preference statements under a *ceteris paribus* (all else equal) assumption. The semantics of our local preference statements capture the classical notion of (conditional) *preferential independence* [13], while our *CP-network* (conditional preference network) representation allows these statements to be organized in a precise way. We also describe several inference algorithms for dominance queries, and show that these are very efficient for certain classes of networks, and seem to work well on general network structures.

Our conditional *ceteris paribus* semantics requires that the user specify, for any specific feature $F$ of interest, which other features can impact her preferences for values of $F$. For each instantiation of the relevant features (parents of $F$), the user must specify her preference ordering over values of $F$ conditional on the parents assuming the instantiated values; for instance, $f_1$ may be preferred to $f_2$ when $g_1$ and $h_2$ hold. Such a preference is given a *ceteris paribus* interpretation: $f_1$ is preferred to $f_2$ given $g_1$ and $h_2$ *all else being equal*. In other words, for any fixed instantiation of the remaining features, an outcome where $f_1$ holds is preferred to one where $f_2$ holds (assuming $g_1$ and $h_2$). Such statements are arguably quite natural and appear in several places (e.g., in e-commerce applications). For instance, the product selection service offered by Active Buyer's Guide asks for (unconditional) *ceteris paribus* statements in assessing a user's preference for various products.[1] Conditional expressions offer even greater flexibility. Generally, tools for representing and reasoning about *ceteris paribus* preferences are important because they should aid in the elicitation process for naive users.

---

[1] See www.activebuyersguide.com. The tools there also ask for some semi-quantitative information about preferences.

Preference elicitation is a complex task and is a key focus in work on decision analysis [13, 11, 9], especially elicitation involving expert users. Automating the process of preference extraction can be very difficult. Straightforward approaches involving the direct comparison of all pairs of outcomes are generally infeasible for a number of reasons, including the exponential number of outcomes (in the number of relevant features for which preferences are indicated) and the complexity of the questions that are asked (the comparison of complete outcomes). There has been considerable work on exploiting the structure of preferences and utility functions in a way that allows them to be appropriately decomposed [13, 1]. For instance, if certain attributes are preferentially independent of others [13], one can assign degrees of preference to these attribute values without worrying about other attribute values. Furthermore, if one assumes more stringent conditions, often one can construct an additive value function in which each attribute contributes to overall preference to a certain "degree" (the *weight* of that attribute) [13]. For instance, it is common in engineering design problems to make such assumptions and simply require users to assess the weights [6]. This allows the direct tradeoffs between values of different attributes to be assessed concisely. Case-based approaches have also recently been considered [10].

Models such as these make the preference elicitation process easier by imposing specific requirements on the form of the utility or preference function. We consider our CP-network representation to offer an appropriate tradeoff between allowing flexible preference expression and imposing a particular preference structure. Specifically, unlike much of the work cited above, conditional preference statements will be permitted.

The remainder of the paper is organized as follows. In Section 2, we describe the necessary background on preference functions. We define our graphical preference model, *CP-networks*, in Section 3 and describe its semantics in terms of *ceteris paribus* (*conditional preferential independence*) statements. Though the CP-semantics of the local preference statements could be considered somewhat weak, some surprisingly strong conclusions regarding dominance can often be drawn based on the network structure. In Section 4, we consider the task of answering dominance queries as a search for a sequence of more preferred (or less preferred) alternatives leading to the potentially dominating (or dominated) outcome. We formally define the search space and describe several completeness-preserving pruning techniques. In Section 5, we describe several search strategies, heuristics designed to work effectively for certain types of problems. We show that these heuristics are backtrack-free for certain types of networks and where backtrack points arise for other types. We also describe how to view this problem as a planning problem. To conclude, in Section 6 we briefly describe the use of CP-nets in two abstract applications. The first is the sorting of a product database using the preferences over product features, allowing the most preferred products to be identified for a consumer. The second is the use of CP-nets in constraint-based optimization. Finally, we offer some thoughts on future research.

## 2    Preference Relations

We focus our attention on single-stage decision problems with complete information, ignoring in this paper any issues that arise in multi-stage, sequential decision analysis and any considerations of risk that arise in the context of uncertainty.[2] We begin with an outline of the relevant notions from decision theory. We assume that the world can be in one of a number of *states* $S$ and at each state $s$ there are a number of *actions* $\mathcal{A}_s$ that can be performed. Each action, when performed at a state, has a specific *outcome* (we do not concern ourselves with uncertainty in action effects or knowledge of the state). The set of all outcomes is denoted $\mathcal{O}$. A *preference ranking* is a total preorder $\succeq$ over the set of outcomes: $o_1 \succeq o_2$ means that outcome $o_1$ is equally or more preferred to the decision maker than $o_2$. The aim of decision making under certainty is, given knowledge of a specific state, to choose the action that has the most preferred outcome. We note that the ordering $\succeq$ will be different for different decision makers. For instance, two different customers might have radically different preferences for different types of computer systems that a sales program is helping them configure.

Often, for a state $s$, certain outcomes in $\mathcal{O}$ cannot result from any action $a \in \mathcal{A}_s$: those outcomes that can obtain are called *feasible outcomes* (given $s$). In many instances, the mapping from states and actions to outcomes can be quite complex. In other decision scenarios, actions and outcomes may be equated: a user is allowed to directly select a feasible outcome (e.g., select a product with a desirable combination of features). Often states may play no role (i.e., there is a single state).

What makes the decision problem difficult is the fact that outcomes of actions and preferences are not usually represented so directly. We focus here on preferences. We assume a set of *features* (or variables or attributes) $F = \{F_1, \cdots F_n\}$ over which the decision maker has preferences. Each feature $F_i$ is associated with a domain of *feature values* $\mathcal{F}_i = \{f_1^i, \cdots f_{n_i}^i\}$ it can take. The product space $\mathcal{F} = \mathcal{F}_1 \times \cdots \times \mathcal{F}_n$ is the set of outcomes. Thus direct assessment of a preference function is usually infeasible due to the exponential size of $\mathcal{F}$. We denote a particular assignment of values to a set $X \subseteq F$ as $\vec{x}$, and the concatenation of two such partial assignments to $X$ and $Y$ ($X \cap Y = \emptyset$) by $\vec{x}\vec{y}$. If $X \cup Y = F$, $\vec{x}\vec{y}$ is a (complete) outcome.

Fortunately, a preference function can be specified (or partially specified) concisely if it exhibits sufficient structure.

We describe certain types of structure here, referring to [13] for a detailed description of these (and other) structural forms and a discussion of their implications. These notions are standard in multi-attribute utility theory. A set of features $X$ is *preferentially independent* of its complement $Y = F - X$ iff, for all $\vec{x}_1, \vec{x}_2, \vec{y}_1, \vec{y}_2$, we have

$$\vec{x}_1 \vec{y}_1 \succeq \vec{x}_2 \vec{y}_1 \ \text{ iff } \ \vec{x}_1 \vec{y}_2 \succeq \vec{x}_2 \vec{y}_2$$

In other words, the structure of the preference relation over assignments to $X$, when all other features are held fixed, is the same no matter what values these other features take. If the relation above holds, we say $\vec{x}_1$ is preferred to $\vec{x}_2$ *ceteris paribus*. Thus, one can assess the relative preferences over assignments to $X$ once, knowing these preferences do not change as other attributes vary. We can define conditional preferential independence analogously. Let $X$, $Y$ and $Z$ partition $F$ (each set is nonempty). $X$ and $Y$ are *conditionally preferentially independent* given $\vec{z}$ iff, for all $\vec{x}_1, \vec{x}_2, \vec{y}_1, \vec{y}_2$, we have

$$\vec{x}_1 \vec{y}_1 \vec{z} \succeq \vec{x}_2 \vec{y}_1 \vec{z} \ \text{ iff } \ \vec{x}_1 \vec{y}_2 \vec{z} \succeq \vec{x}_2 \vec{y}_2 \vec{z}$$

In other words, the preferential independence of $X$ and $Y$ only holds when $Z$ is assigned $\vec{z}$. If this relation holds for all assignments $\vec{z}$, we say $X$ and $Y$ are *conditionally preferentially independent* given $Z$.

This decomposability of a preference functions often allows one to identify the most preferred outcomes rather readily. Unfortunately, the *ceteris paribus* component of these definitions ensures that the statements one makes are relatively weak. In particular, they do not imply a stance on specific value tradeoffs. For instance, suppose two features $A$ and $B$ are preferentially independent so that the preferences for values of $A$ and $B$ can be assessed separately; e.g., suppose $a_1 \succ a_2$ and $b_1 \succ b_2$. Clearly, $a_1 b_1$ is the most preferred outcome and $a_2 b_2$ is the least; but if feasibility constraints make $a_1 b_1$ impossible, we must be satisfied with one of $a_1 b_2$ or $a_2 b_1$. We cannot tell which is most preferred using these separate assessments. However, under stronger conditions (e.g., *mutual preferential independence*) one can construct an additive value function in which weights are assigned to different attributes (or attribute groups). This is especially appropriate when attributes take on numerical values. We refer to [13] for a discussion of this problem.

Given such a specification of preferences, a number of different techniques can be used to search the space of feasible outcomes for a most preferred outcome.

## 3    CP-Networks

In this section we describe a network representation that allows the compact (but generally incomplete) representation of a preference relation. We first describe the basic model and its semantics and then describe inference procedures for dominance testing.

Our representation for preferences is graphical in nature, and exploits conditional preferential independence in structuring a user's preferences. The model is similar to a Bayes net on the surface; however, the nature of the relation between nodes within a network is generally quite weak (e.g., compared with the probabilistic relation in Bayes nets). Others have defined graphical representations of preference relations; for instance Bacchus and Grove [1] have shown some strong results pertaining to undirected graphical representations of additive independence. Our representation and semantics is rather distinct, and our main aim in using the graph is to capture statements of conditional preferential independence. We note that reasoning about *ceteris paribus* statements has been explored in AI, though not in the context of network representations [7].

For each feature $F$, we ask the user to identify a set of *parent* features $P(F)$ that can affect her preference over various $F$ values. That is, given a particular value assignment to $P(F)$, the user should be able to determine a preference order for the values of $F$, all other things being equal. Formally, denoting all other features aside from $F$ and $P(F)$ by $\overline{F}$, we have that $F$ and $\overline{F}$ are conditionally preferentially independent given $P(F)$. Given this information, we ask the user to explicitly specify her preferences over $F$ values for all possible $P(F)$ values. We use the above information to create an annotated graph in which each feature $F$ has $P(F)$ as its set of parents. The node $F$ is annotated with a *condition preference table* (CPT) describing the user's preferences over $F$'s values given every combination of parent values.[3] We call these structures *conditional preference networks* (or CP-networks). We note that nothing in the semantics forces the graph to be acyclic, though we argue below that most natural networks will indeed be acyclic. Moreover, even cyclic CP-networks cannot express all possible total preference orderings, as can be shown by a simple counting argument.

We illustrate the network semantics and some of its consequences with a series of examples. In the following examples all features are boolean, though our semantics is defined for features with arbitrary finite domains.

**Example 1**  Asking the user to describe her preference over feature $B$, we are told that this preference depends on the value for $A$ and on that value alone (*ceteris paribus*). We then make $A$ a parent of $B$ and ask about her preference on $B$ for each value of $A$. She may say that, when $a$ holds, she prefers $b$ over $\overline{b}$, and when $\overline{a}$ holds she prefers $\overline{b}$ over $b$, *ceteris paribus*. This is written as:

$$a : b \succ \overline{b}$$
$$\overline{a} : \overline{b} \succ b$$

---

[3] That is, we assume that a preorder is provided over the domain of $F$, such that for any two values $f_i$ and $f_j$, either $f_i \succ f_j$, $f_j \succ f_i$, or $f_i$ and $f_j$ are equally preferred. For ease of presentation, we ignore indifference in our algorithms (though its treatment is straightforward). We assume this relation is fully specified (though see Section 6).

**Example 2** Suppose we have two features $A$ and $B$, where $A$ is a parent of $B$ and $A$ has no parents. Assume the following conditional preferences:

$$a \succ \overline{a}; \quad a : b \succ \overline{b}; \quad \overline{a} : \overline{b} \succ b$$

Somewhat surprisingly, this information is sufficient to totally order the outcomes:

$$ab \succ a\overline{b} \succ \overline{a}\overline{b} \succ \overline{a}b.$$

Notice that we can judge each outcome in terms of the conditional preferences it violates. The $ab$ outcome violates none of the preference constraints. Outcome $a\overline{b}$ violates the conditional preference for $B$. Outcome $\overline{a}b$ violates the preference for $A$. Outcome $\overline{a}b$ violates both. What is surprising is that the *ceteris paribus* semantics implies that violating the $A$ constraint is worse than violating the $B$ constraint (we have $a\overline{b} \succ \overline{a}\overline{b}$). That is, the parent preferences have higher priority than the child preferences.

**Example 3** Suppose we have three features $A$, $B$, and $C$, and suppose that the preference dependency graph is disconnected. Let's assume that $a \succ \overline{a}$, $b \succ \overline{b}$, and $c \succ \overline{c}$. Given this information we can conclude that $abc$ is the most preferred outcome, then comes $\overline{a}bc$, $a\overline{b}c$, and $ab\overline{c}$. These three cannot be ordered based on the information provided. Less preferred than the last two is $a\overline{b}\overline{c}$, and so on. The least preferred outcome is $\overline{a}\overline{b}\overline{c}$.

**Example 4** Suppose we have three features $A$, $B$, and $C$, and the conditional preference graph forms a chain with $A$ having no parents, $A$ the parent of $B$, and $B$ the parent of $C$. Suppose we have the following dependence information:

$$a \succ \overline{a}; \quad a : b \succ \overline{b}; \quad \overline{a} : \overline{b} \succ b; \quad b : c \succ \overline{c}; \quad \overline{b} : \overline{c} \succ c$$

These preference constraints imply the following ordering:

$$abc \succ ab\overline{c} \succ a\overline{b}\overline{c} \succ a\overline{b}c \succ \overline{a}\overline{b}c \succ \overline{a}\overline{b}\overline{c} \succ \overline{a}b\overline{c},$$

which totally orders all but one of the outcomes. Notice how we get from one outcome to the next in the chain: we *flip* (or exchange) the value of exactly one feature according to the preference dependency information. The element not in this chain is $\overline{a}b\overline{c}$, and we can derive the ordering $a\overline{b}\overline{c} \succ \overline{a}b\overline{c} \succ \overline{a}b\overline{c}$. Thus, the only two outcomes not totally ordered are $\overline{a}b\overline{c}$ and $a\overline{b}c$. From Example 2, we saw that violations of preference constraints for parent features are worse than violations of constraints over child preferences. In one of the two unordered outcomes we violate the preference of the most important feature ($A$), while in the other outcome we violate preference over two less important features ($B$ and $C$). The semantics of CP-networks does not specify which of these tuples is preferred.

There are two important things to notice about these examples. First, a chain of "flipping feature values" can be used to show that one outcome is better than another. In Example 4, the conditional preferences for $C$ allow the value of

$C$ to be "flipped" in outcome $\overline{a}b\overline{c}$ to obtain $\overline{a}bc$. $B$'s value can then be flipped (given $\overline{a}$) to obtain $\overline{a}\overline{b}c$, and so on. Second, violations are worse (i.e., have a larger negative impact on preference) the higher up they are in the network, although we cannot compare two (or more) lower level violations to violation of a single ancestor constraint. These observations underly the inference algorithms below.

As mentioned, the semantics of CP-nets do not preclude cyclic networks. For instance, a two-variable network where $A$ depends on $B$ and $B$ depends on could be consistently quantified as follows:

$$a : b \succ \overline{b}; \quad \overline{a} : \overline{b} \succ b$$
$$b : a \succ \overline{a}; \quad \overline{b} : \overline{a} \succ a$$

Under these preferences, the user simply prefers $A$ and $B$ to have the same value, with both $ab$ and $\overline{a}\overline{b}$ maximally preferred. Acyclic graphs always have a unique most-preferred outcome. We note that cyclic preference graphs can be inconsistent (e.g., in the example above, simply reverse the conditional preferences for $B$ under each value of $A$). Indeed, acyclic graphs are always consistent (i.e., correspond to at least one well-defined preference ordering). It seems there is rarely a need for cyclic structures unless one wants to express indifference between certain assignments to subsets of variables. In this case, one can often cluster the variables to maintain acyclicity. In what follows, we assume that our CP-nets are acyclic.

## 4 Searching for Flipping Sequences

We assume we are given an acyclic CP-network over features $F_1, \cdots F_n$. By convention, we assume the ordering of these features respects the topology of the network (that is, the parents of any $F_i$ have indices $j < i$). We use $x_i, x_i', y_i$, etc. to denote values of feature $F_i$. The basic inference problem we address is the following: given a CP-network $N$, and two outcomes $\mathbf{x} = x_1 x_2 \cdots x_n$, $\mathbf{y} = y_1 y_2 \cdots y_n$, is $\mathbf{x} \succ \mathbf{y}$ a consequence of preferences of the CP-network? In other words, is the outcome $\mathbf{x}$ preferred to $\mathbf{y}$? We treat the inference problem as a search for a *flipping sequence* from the (purported) less preferred outcome $\mathbf{y}$, through a series of more preferred outcomes, to the (purported) more preferred outcome $\mathbf{x}$, where each value flip in the sequence is sanctioned by the network $N$. Conversely, we can view the problem as a search in the opposite direction, from the more preferred outcome to the less preferred outcome.

### 4.1 Improving Search

Given any CP-network, and a query $\mathbf{x} \succ \mathbf{y}$, we define the *improving search tree* as follows. The search tree is rooted at $\mathbf{y} = y_1 y_2 \cdots y_n$; the children of any node $\mathbf{z} = z_1 z_2 \cdots z_n$ in the search tree are those outcomes that can be reached by changing one feature value $z_i$ to $z_i'$ such that $z_i' \succ z_i$ given the values $z_j, j < i$. Note that possible improving values $z_i'$ of $F_i$ can be readily determined by inspecting the CPT for $F_i$. Since the only preference statements explicitly represented in the network are those captured by the CPTs, it is clear that $x \succ y$ is implied by $N$ iff there exists a path from $y$ to $x$ in the improving search tree. Thus, any complete search procedure—any procedure guaranteed to examine every branch of the search tree—will be a sound and complete query answering procedure. All procedures discussed in this paper are, in this sense, sound and complete.

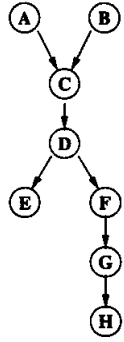

Figure 1: An Example Conditional Preference Graph

**Example 5** Consider the preference graph of Figure 1. Suppose that the conditional preferences are:

$$a \succ \overline{a}; \quad b \succ \overline{b};$$
$$(a \wedge b) \vee (\overline{a} \wedge \overline{b}) : c \succ \overline{c}; \quad (a \wedge \overline{b}) \vee (\overline{a} \wedge b) : \overline{c} \succ c$$
$$c : d \succ \overline{d}; \quad \overline{c} : \overline{d} \succ d; \quad d : e \succ \overline{e}; \quad \overline{d} : \overline{e} \succ e;$$
$$d : f \succ \overline{f}; \quad \overline{d} : \overline{f} \succ f; \quad f : g \succ \overline{g}; \quad \overline{f} : \overline{g} \succ g$$
$$g : h \succ \overline{h}; \quad \overline{g} : \overline{h} \succ h$$

Suppose we want to compare outcome $abcde\overline{f}\overline{g}h$ (which violates the $G$ preference) and outcome $\overline{a}\overline{b}c\overline{d}\overline{e}f\overline{g}h$ (which violates the $A$ preference). In order to show that the first is preferred, we generate the sequence: $\overline{a}\overline{b}c\overline{d}\overline{e}f\overline{g}h \prec \overline{a}b\overline{c}\overline{d}\overline{e}f\overline{g}h \prec abc\overline{d}\overline{e}f\overline{g}h \prec abcd\overline{e}f\overline{g}h \prec abcde\overline{f}\overline{g}h$. Intuitively, we constructed a sequence of increasingly preferred outcomes, using only valid conditional independence relations represented in the CP-network, by *flipping* values of features. We are allowed to change the value of a "higher priority" feature (higher in the network) to its preferred value, even if this introduces a new preference violation for some lower priority feature (a descendent in the network). For instance, the first flip of $A$'s value in this sequence to its preferred state repairs the violation of $A$'s preference constraint, while introducing a preference violation with respect to $C$ (the value $\overline{c}$ is dispreferred when $ab$ holds). This process is repeated (e.g., making $C$ take its conditionally most preferred value at the expense of violating the preference for $D$) until the single preference violation of $F$ (in the "target" outcome) is shown to be preferred to the single preference violation of $A$ (in the initial outcome). This demonstrates how the violation of conditional preference for a feature is dispreferred to the violation of one of its descendent's preferences.

Suppose we compare $abcde\overline{f}\overline{g}h$ (which violates the $G$ preference and the $H$ preference) and $\overline{a}\overline{b}c\overline{d}\overline{e}f\overline{g}h$ (which violates the $A$ preference). These turn out not to be comparable (neither is preferred to the other). The sequence of flips above cannot be extended to change the values of both $G$ and $H$ so that their preference constraints are violated. The sole violation of the $A$ constraint cannot be dominated by the violation of two (or more) descendents *in a chain*.

If we want to compare $abcd\overline{e}f\overline{g}h$ (which violates the $E$ preference and the $G$ preference) and $\overline{a}\overline{b}c\overline{d}\overline{e}f\overline{g}h$ (which violates the $A$ preference), we can use the following sequence: $\overline{a}\overline{b}c\overline{d}\overline{e}f\overline{g}h \prec ab\overline{c}\overline{d}\overline{e}f\overline{g}h \prec abc\overline{d}\overline{e}f\overline{g}h \prec abcd\overline{e}f\overline{g}h$. The violation of $E$ and $G$ is preferred to the violation of $A$: intuitively, the $A$ violation can be absorbed by violation in *each path starting at $D$*.

Now consider the comparison of $abcdefgh$ (which violates the $G$ and $H$ preferences) and $\overline{a}\overline{b}cdefgh$ (which violates the $A$ and $B$ preferences). We can use the following sequence of flips to show preference: $\overline{a}\overline{b}cdefgh \prec \overline{a}b\overline{c}defgh \prec a\overline{b}c\overline{d}efgh \prec ab\overline{c}\overline{d}efgh \prec a\overline{b}c\overline{d}\overline{e}fgh \prec abc\overline{d}\overline{e}fgh \prec abcd\overline{e}fgh \prec abcd\overline{e}\overline{f}gh \prec abcde\overline{f}gh \prec abcdefgh$. This shows how two violations in ancestor features covers two violations in their descendents.

These examples illustrate how certain preference violations have priority over others in determining the relative ordering of two outcomes. Intuitively, dominance is shown by constructing a sequence of legal flips from the initial outcome to the target.

### 4.2 Worsening Search

A query $x \succ y$ can also be answered using search through the *worsening search tree*, defined as follows. The search tree is rooted at $x = x_1 x_2 \cdots x_n$; the children of any node $z = z_1 z_2 \cdots z_n$ in the search tree are those outcomes that can be reached by changing one feature value $z_i$ to $z_i'$ such that $z_i' \prec z_i$ given the values $z_j, j < i$. Note that possible worsening values $z_i'$ of $F_i$ can be readily determined by inspecting the CPT for $F_i$. Again, it is clear that $x \succ y$ is implied by $N$ iff there exists a path from $x$ to $y$ in the worsening search tree.

While clearly any path from $x$ to $y$ that exists in the worsening search tree corresponds to a path from $y$ to $x$ in the improving search tree, and vice versa, the search space may be such that searching in the improving search tree is most effective for some queries, while searching in the worsening search tree is most appropriate for others.

**Example 6** Consider the CP-network described in Example 4. Suppose we wish to test whether $a\overline{b}c \succ \overline{a}b\overline{c}$. Taking $a\overline{b}c$ as the root of the worsening tree, the only path one can generate is $a\overline{b}c \succ \overline{a}\overline{b}c \succ \overline{a}bc \succ \overline{a}b\overline{c}$. In other words, the worsening tree does not branch and leads directly to a positive answer to the query. In contrast, the improving search tree rooted at $\overline{a}b\overline{c}$ consists of six branches (with a maximum length of seven nodes), and only one path leads to a solution (see Figure 2).

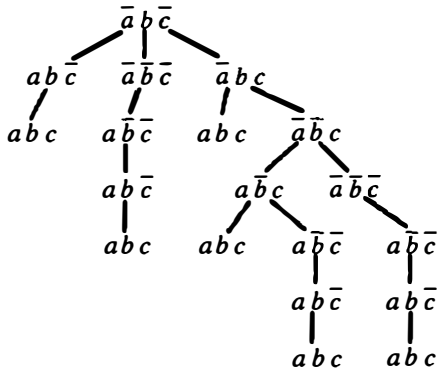

Figure 2: Improving Search Tree from $\bar{a}b\bar{c}$ (Example 6)

**Example 7** With the same network, consider the query $abc \succ \bar{a}\bar{b}\bar{c}$. Taking $\bar{a}\bar{b}\bar{c}$ as the root of the improving search tree, the only path in the tree is $\bar{a}\bar{b}\bar{c} \prec a\bar{b}\bar{c} \prec ab\bar{c} \prec abc$. In contrast, the worsening search tree rooted at $abc$ consists of six branches (with a maximum length of seven nodes), and only one path leads to a solution.

For this reason, we believe that a parallel search in both the improving and worsening search trees is generally most appropriate. Though we have illustrated positive queries only, the same considerations apply to negative queries, where, in fact, exploiting small search trees is especially important in order to quickly fail.

### 4.3 Suffix Fixing and Extension

Though we haven't yet detailed specific search procedures, in the remainder of this section we suppose that we have some complete (and necessarily sound) search procedure. Regardless of whether one uses improving or worsening search, there are two simple rules that allow one to make deterministic moves in search space (i.e., choose flips that need not be backtracked over, or reconsidered) without impacting completeness of the search procedure.

The first rule is *suffix fixing*. We define a *suffix* of an alternative $z = z_1 z_2 \cdots z_n$ to be some subset of the values $z_i z_{i+1} \cdots z_n$, $i \geq 1$. A suffix can be defined for any legal ordering of the features. Suppose an improving search for the query $x \succ y$ takes us from the root node $y$ to node $z = z_1 z_2 \cdots z_n$. Suppose further that some *suffix* of $z$ matches the suffix of target $x$; that is, $z_j = x_j$ for all $j \geq i$.[4] The suffix fixing rule requires that those features making up the suffix never be flipped. The following proposition ensures that we never need reconsider a decision not to flip features in a matching suffix.

**Proposition 1** *Let there be a path in the improving search*

---

[4] The matching suffix can be "created" by a reordering of the features that is consistent with the partial ordering of the (acyclic) CP-network.

*tree from root $y$ to node $z$, such that some suffix of $z$ matches that of the target $x$. If there is a path from $y$ to $x$ that passes through $z$, then there is a path from $z$ to $x$ such that every node along that path has the same values as $z$ for the features that make up the suffix.*

This effectively restricts the search tree under $z$ to have only paths that retain the suffix values. Though one may have to backtrack over choices that lead to $z$, one will not have to consider the full search tree under $z$. The suffix fixing rule also applies to worsening search.

A second completeness-preserving rule is the *suffix extension rule*. Suppose that a path to intermediate node $z$ has been found that matches some suffix of the target $x$. Furthermore, suppose that the values of $z$ allow this suffix to be extended; that is, suffix $z_i z_{i+1} \cdots z_n$ matches the target and feature $F^{i-1}$ can be improved from $z_{i-1}$ to $z'_{i-1} = x_{i-1}$.[5] Then the flip to $z'_{i-1}$ can be chosen and not reconsidered.

**Proposition 2** *Let there exist a path in the improving search tree from root $y$ to node $z$, such that some suffix of $z$ matches that of the target $x$, and that the suffix can be extended by a legal move from $z$ to $z'$. If there exists a path from $y$ to $x$ that passes through $z$, then there exists a path from $z'$ to $x$ such that every node along that path has the same values as $z'$ for the features that make up the extended suffix.*

**Example 8** Consider the CP-network of Figure 1 with the conditional preferences as in Example 5. Suppose we were to consider the query

$$abcd\bar{e}\bar{f}\bar{g}h \succ \bar{a}b\bar{c}\bar{d}\bar{e}fgh$$

using an improving search. Suffix fixing means that we never have to consider flipping $g$, $h$ or $e$ (there is a reordering of the features that has these three as the rightmost features). The suffix extension rule means that we can flip $f$ to $\bar{f}$ (as $\bar{d} : \bar{f} \succ f$), without backtracking over this choice. We cannot immediately flip $\bar{d}$ to $d$ in the context of $\bar{c}$, so suffix extension is not applicable (once $f$ is flipped).

### 4.4 Forward Pruning

In this section we describe a general pruning mechanism that can be carried out given a query $x \succ y$. It
- often quickly shows that no flipping sequence is possible;
- prunes the domains of the features to reduce the flipping search space;
- doesn't compromise soundness or completeness; and
- is relatively cheap (time is $O(nrd^2)$ where $n$ is the number of features, $r$ is the maximum number of conditional preference rules for each feature, and $d$ is the size of the biggest domain).

The general idea is to sweep forward through the network, pruning any values of a feature that cannot appear in any (improving or worsening) flipping sequence to validate a query. Intuitively, we consider the set of flips possible, ignoring interdependence of the parents and the number of times the parents can change their values.

---

[5] Again, the suffix can be found using feature reordering.

We consider each feature in an order consistent with the network topology (so that parents of a node are considered before the node). For each feature $F$, we build a graph with nodes corresponding to the possible values for $F$, and for each conditional preference relation

$$c : v_1 \succ v_2 \succ \cdots \succ v_d$$

such that $c$ is consistent with the pruned values of the parents of $F$, we include an arc between the successive values (i.e., between the values $v_i$ and $v_{i+1}$).

We can prune any value that isn't on a directed path from x's value for feature $F$ to y's value for feature $F$. This can be implemented by running Dijkstra's algorithm [5] twice: once to find the nodes reachable from x's value for feature $F$ and again to find the nodes that can reach y's value for feature $F$. These sets of nodes can be intersected to find the possible values for $F$. If there are no nodes remaining, the domination query fails: there is no legal flipping sequence. This often results in quick failure for straightforward cases, so that we only carry out the search for non-obvious cases.

**Example 9** Consider the CP-network of Figure 1 with the conditional preferences as in Example 5. Consider a query of the form

$$a\overline{b} \ldots \succ \overline{a}b \ldots$$

First we consider $A$. We can draw an arc $a \rightarrow \overline{a}$, and find that both $a$ and $\overline{a}$ are on a path, so no values of $A$ are pruned.[6] We then consider $B$ and draw an arc $b \rightarrow \overline{b}$; but there are no paths from $\overline{b}$ to $b$, so the query fails quickly without looking at the other features.

One could imagine extending this pruning phase to include more information, such as the sequences of values through which the parents can pass. From this one can determine the possible sequences of values through which the child feature could pass. Generally, the combinatorics of maintaining such sequences is prohibitive; but in the binary case, any path through the set of values is completely determined by the starting value and a count of the number of times the value flips. Pruning still ignores the possible interdependencies of the values for the parents, but for singly-connected networks (where we can guarantee the sequences of values the parents can pass through are independent), pruning is complete in the sense that if it stops without failing there is a flipping sequence. This was the basis of the counting algorithm in [3] for singly-connected binary CP-networks.

---

[6]If the example were changed slightly so that $A$ had a third value $\overline{\overline{a}}$, where $a \succ \overline{a} \succ \overline{\overline{a}}$, then this third value could be pruned from $A$, thus simplifying the tables for all the children of $A$.

## 5    Search Strategies and Heuristics

In the previous section, the search space was formally defined, and several completeness preserving rules for pruning the search space were defined. This leaves open the issue of effective procedures for searching. In this section we describe several heuristics for exploring the search tree. We first describe some simple heuristics that seem to be effective for many networks, and are, in fact, backtrack-free for certain classes of networks. We then show how this search problem can be recast as a planning problem and briefly describe the potential benefits of such a view.

### 5.1    Rightmost and Least-Improving Heuristics

The rightmost heuristic requires that the variable whose value one flips when deciding which child to move to is the rightmost variable that can legally be flipped. For instance, consider the improving search tree in Example 6 (as illustrated in Figure 2). Given a target outcome $\overline{a}\overline{b}c$, we see that the rightmost heuristic leads us directly to the target in two steps. If the target outcome were different, say $abc$, then the rightmost heuristic has the potential to lead us astray. However, when we incorporate the suffix-fixing rule into the search, we see that the rightmost heuristic (defined now as flipping the rightmost value that doesn't destroy a suffix match) will lead directly to *any* target outcome in the search tree. For example, given target outcome $abc$, the rightmost heuristic discovers the shortest path to the target: notice also that suffix-fixing prevents us from exploring the longest (length six) path to the target.

This example suggests that for chains, the rightmost heuristic will lead to a proof, if one exists, without backtracking. This may not be the the case, however, if variables are not all binary.

**Example 10** Consider the CP-network where variable $A$, with domain $\{a_1, a_2, a_3\}$, is a parent of boolean variable $B$. Conditional preferences are given by

$$a_1 \succ a_2 \succ a_3$$
$$a_1 : b \succ \overline{b}; \quad a_2 : \overline{b} \succ b; \quad a_3 : b \succ \overline{b}$$

Given query $a_1\overline{b} \succ a_3b$, the rightmost heuristic in an improving search could first construct the sequence $a_3b \prec a_1b$, reaching a dead end (thus requiring backtracking). The direct sequence $a_3b \prec a_2b \prec a_2\overline{b} \prec a_1\overline{b}$ is also consistent with the rightmost heuristic.

In the example above, the rightmost heuristic permitted a "jump" from $a_3$ to the most preferred value $a_1$ without moving through the intermediate value $a_2$. This prevented it from discovering the correct flipping sequence.

In multivalued domains, another useful heuristic is the *least improving heuristic* (or in worsening searches, the *least worsening heuristic*): when the rightmost value can be flipped to several improving values given its parents, the improving value that is *least preferred* is adopted. This allows greater flexibility in the movement of "downstream" variables. While one can always further improve the value of the variable in question from its least improving value to a more preferred value (provided that parent values are maintained), "skipping" values may prevent us from setting its descendents to their desired values.

Both the rightmost and least improving heuristics can be viewed as embodying a form of *least commitment*. Flipping the values of the rightmost possible variable (i.e., a variable with the smallest number of descendents in the network) can be seen as leaving maximum flexibility in flipping the values of other variables. An upstream variable limits the possible flipping sequences more drastically than a downstream variable—specifically, altering a specific variable does not limit the ability to flip the values of its nondescendents. For the reasons described above, the least improving heuristic can be cast in a similar light.

Unfortunately, while the least-commitment approach works well in practice, it does not allow backtrack-free search in general, as the following example shows.

**Example 11** Consider the CP-network with three variables $A$, $B$ and $C$ such that $A$ is the only parent of $B$ and $B$ is the only parent of $C$. Suppose $A$ has domain $\{a, \overline{a}\}$, $B$ has domain $\{b_1, b_2, b_3\}$ and $C$ has domain $\{c, \overline{c}\}$, with the following conditional preferences:

$$a \succ \overline{a};$$
$$a : b_3 \succ b_2 \succ b_1;$$
$$\overline{a} : b_3 \succ b_1 \succ b_2;$$
$$b_2 : \overline{c} \succ c; \qquad b_1 \vee b_3 : c \succ \overline{c}$$

Consider the query $ab_3\overline{c} \succ \overline{a}b_1c$ with an improving search. $c$ cannot be improved in the context of $b_1$. However $b_1$ can be improved to $b_3$ in the context of $\overline{a}$, but this leads to a dead end. The right thing to do is to flip $a$ first, then change $b_1$ to $b_2$ which will let you flip $c$ and then change $b_2$ to $b_3$.

While queries over chain-structured networks with multivalued variables cannot reliably be searched backtrack-free using the rightmost and least-improving heuristics, this search approach is backtrack-free for chains when all variables are binary. Intuitively, this is the case because changing the value of the rightmost allowable variable does not impact the ability to flip its parent's value; furthermore, changing this variable cannot prevent its child from being flipped, since if the child needed a different value (and could have been flipped), it would have been flipped earlier. For similar reasons, binary tree-structured networks (where every variable has at most one parent, but perhaps multiple children) can also be searched backtrack-free.

**Example 12** Consider the binary tree-structured preference graph of Figure 3 with the conditional preferences:

$$a \succ \overline{a}$$
$$a : b \succ \overline{b}; \quad \overline{a} : \overline{b} \succ b$$
$$a : c \succ \overline{c}; \quad \overline{a} : \overline{c} \succ c$$

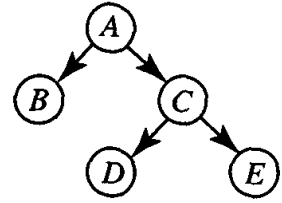

Figure 3: A Tree-Structured Conditional Preference Graph

$$c : d \succ \overline{d}; \quad \overline{c} : \overline{d} \succ d$$
$$c : e \succ \overline{e}; \quad \overline{c} : \overline{e} \succ e$$

Consider the query $a\overline{b}\overline{c}\overline{d}e \succ \overline{a}bcde$. Suppose we are searching for an improving flipping sequence from $\overline{a}bcde$. By suffix fixing, we leave $e$ untouched. The first value we flip is $c$. Since this is the only way we could ever get to flip $d$, and because $c$ is binary, there is only ever one other value it could have. We can now flip $d$ forming $\overline{a}b\overline{c}de$ ($\overline{d}$ and $e$ then remain untouched). We can flip $b$, and fix it by suffix fixing (as there is an ordering where it is part of the fixed suffix). The only value we can flip at this point is $a$; this gives us $a\overline{b}\overline{c}de$. We can now flip $c$ and we are done.

**Proposition 3** *The rightmost search heuristic, in conjunction with suffix-fixing and suffix-extension, is complete and backtrack-free for chain- and tree-structured CP-nets with binary variables.*

Polytrees (singly-connected networks containing no undirected cycles) cannot be searched without backtracking in general, even when variables are binary. This is due to the fact that several parents of a given node may each be allowed to have their values flipped, but only one of the choices may lead to the target outcome, while the others lead to deadends. For instance, suppose we consider Example 5, restricted to the variables $A, B, C$, and are given the query $a\overline{b}\overline{c} \succ \overline{a}bc$. Using an improving search rooted at $\overline{a}bc$, we have a choice of flipping $A$ or $B$. If $B$ is chosen, we start down the path $\overline{a}\overline{b}c \prec \overline{a}bc$; but this clearly cannot lead to the target, since there is no way to flip $B$ back to $\overline{b}$. A deadend will be reached and we must backtrack to flip $A$ before $B$, leading to the solution path $\overline{a}\overline{b}c \prec a\overline{b}c \prec a\overline{b}\overline{c}$.

Essentially, this means we have to may have to consider different variable orderings over the ancestors of a given node. It turns out that these are the only backtrack points in binary polytrees.

Finally, for general (multiply-connected) CP-nets, complex interdependencies can exist among the parents of variables because the parents themselves may share ancestors. This can lead to complex search paths in the successful search for a flipping sequence. Though we don't provide examples, one can construct networks and specific queries such that no *fixed* ordering of variables allows the rightmost heuristic to work backtrack-free. We also note that the shortest flipping sequence for certain queries can be exponential in length given a maximally-connected acyclic network (e.g., we can require sequences of length $O(2^{n/2})$ in an $n$-variable binary network). We do not believe such sequences are required in singly-connected networks.

It should be noted that while one can generate example networks and queries that require complicated search, involving considerable backtracking using most simple heuristics, such examples tend to be rather intricate and obscure. They invariably require a tight interaction between the network structure, the conditional preference statements quantifying the network, and the specific query itself. None of the natural examples we have seen require much search.

### 5.2 Flipping Sequences as Plans

In this paper we have considered searching directly for flipping sequences. This can be seen as a case of state-space search. It is also possible to think about answering dominance queries as a type of planning problem. A conditional preference statement of the form

$$c : v_1 \succ v_2 \succ \cdots \succ v_d$$

can be converted into a set of STRIPS actions for improving the value for a variable. In particular, this conditional preference statement can be converted into a set of $d - 1$ STRIPS operators of the form (for $1 < i \leq d$):

**Preconditions:** $c \wedge v_i$

**Add List:** $v_{i-1}$

**Delete list:** $v_i$

This corresponds to the action of improving $v_i$ to $v_{i-1}$ in the context of $c$. (A different set of actions would be created for worsening.)

Given a query $\mathbf{x} \succ \mathbf{y}$, we treat $\mathbf{y}$ as the start state and $\mathbf{x}$ as the goal state. It is readily apparent that that the query is a consequence of the CP-network if and only if there is a plan for the associated planning problem. A plan corresponds to a flipping argument.

The previous algorithms can be viewed as state-based forward planners. It is often the case that domain-specific heuristics can be easily added to a forward search [2], and we expect the same here. We could also use other planning techniques such as regression, partial-order planning, planning as satisfiability and stochastic local search methods for this problem. The application of regression and partial-order planners (more generally, backchaining planners) can provide support for reasoning about the changes in ancestor values required for a specific descendent to flip its value to its target. We note that the planning problems generated by CP-queries will generally look quite different in form from standard AI planning problems, as there are many more actions, and each action is directed toward achieving a particular proposition and requires very specific preconditions.

## 6 Concluding Remarks

In this paper we introduced CP-networks, a new graphical model for representing qualitative preference orderings which reflects conditional dependence and independence of preference statements under a *ceteris paribus* semantics. This formal framework often allows compact and arguably natural representations of preference information. We argued that given a CP-network, the basic inference problem of determining whether one of two given vectors of feature values is preferred to the other is equivalent to the task of finding a connecting sequence of flipping individual feature values. We characterized the corresponding search space and described several strategies and heuristics which often significantly reduce the search effort and allow one to solve many problem instances efficiently.

We see various applications of CP-networks and our dominance testing strategies and heuristics. One of these is sorting a product database according to user-specified preferences. This problem is highly relevant in the context of electronic commerce. Several rather conceptually simplistic implementations are available on the World Wide Web (e.g., Active Buyers Guide). The general idea is to assist a user in selecting a specific product from a database according to her preferences. Here, it is very important to use compact and natural representations for preference information. CP-networks extend current models (which typically don't allow conditional preference statements). Another important aspect of this problem is that the given database precisely defines the items (represented as vectors of feature values) available, and preference information is only required to such an extent that the choice is sufficiently narrowed down to a small selection of products from this database. Dominance testing strategies are important in this context to find a set of Pareto-optimal choices given the (conditional) preference information extracted from the user. Here, an interactive and dynamic approach appears to be most promising, where the user is prompted for additional preference statements until the ordering of the items in the database is sufficiently constrained by the preference information to offer a reasonably small selection of products. While the dominance algorithms are an important part of the database sorting task, the problem does not generally require that all pairwise comparisons be run to completion. Certain preprocessing steps can be taken, that exploit the network structure, to partition tuples in the database according to values of high priority attributes.

Another application area is constraint-based configuration, where the task is to assemble a number of components according to user preferences such that given compatibility constraints are satisfied [3, 6]. A simple example of this is the assembly of components for computer systems where, for instance, the type of system bus constrains the choice of video and sound cards. CP-networks can be used to represent the user preferences which are used together with compatibility constraints to search for most preferred, feasible configurations. In contrast to the database sorting application above, here the set of possible vectors of feature values (i.e., configurations) is not explicitly given, but implicitly specified by the compatibility constraints. Dominance testing is again required for finding most preferred solutions, but now it has to be combined with mechanisms which limit the search to feasible configurations [3].

We are currently extending this work in two directions. First, the search strategies and heuristics for dominance testing presented in this paper have to be implemented in order to empirically assess their performance on various types of problem instances, including real-world problems, as well as handcrafted examples exhibiting uniform, regular structures of theoretical interest. Secondly, we are working on various extensions of the framework presented here. These include cases where the conditional preference statements contain a small amount of quantitative information. In particular, existing applications (such as online interactive consumer guides) suggest that a limited amount of such quantitative preference information might be relatively easy to extract from the user in a natural way, and is very useful for inducing stronger preference orderings.

Another interesting issue is the extension of the representation and reasoning system such that incompletely specified conditional preference information (i.e., incomplete CP-tables) can be taken into account. This is motivated by the fact that often the full preference information given by the CP-tables is not required for deciding a particular dominance query. Therefore, it seems to be useful to consider mechanisms which allow incompletely specified CP-tables and dynamically prompt the user for additional preference information when it is needed.

Finally, we intend to investigate the tradeoffs between the amount of user-interaction required for extracting the preference information and the amount of computation needed for determining most preferred feature vectors. By asking very specific questions about particular, potentially complex preferences, finding most preferred feature vectors can become much easier. On the other hand, asking too many questions, especially those not really necessary for establishing relevant preferences, will annoy the user and make the system less usable. Thus, finding good tradeoffs between the amount of user-interaction and computation time for answering queries—such as finding most preferred items from a database or optimal configurations—seems to be a promising direction for future research. This is related to the motivation underlying goal programming [8, 12]. The representations and search techniques presented in this paper form a starting point for such investigations.

**Acknowledgements:** This research was supported by IRIS-III Project "Interactive Optimization and Preference Elicitation" (BOU).

## References

[1] Fahiem Bacchus and Adam Grove. Graphical models for preference and utility. In *Proceedings of the Eleventh Conference on Uncertainty in Artificial Intelligence*, pages 3–10, Montreal, 1995.

[2] Fabiem Bacchus and Froduald Kabanza. Using temporal logic to control search in a forward chaining planner. In *Proceedings of the 3rd European Workshop on Planning*, 1995. Available via the URL ftp://logos.uwaterloo.ca:/pub/tlplan/tlplan.ps.Z.

[3] Craig Boutilier, Ronen Brafman, Chris Geib, and David Poole. A constraint-based approach to preference elicitation and decision making. In *AAAI Spring Symposium on Qualitative Decision Theory*, Stanford, 1997.

[4] U. Chajewska, L. Getoor, J. Norman, and Y. Shahar. Utility elicitation as a classification problem. In *Proceedings of the Fourteenth Conference on Uncertainty in Artificial Intelligence*, pages 79–88, Madison, WI, 1998.

[5] Thomas H. Cormen, Charles E. Lierson, and Ronald L. Rivest. *Introduction to Algorithms*. MIT Press, Cambridge, MA, 1990.

[6] Joseph G. D'Ambrosio and William P. Birmingham. Preference-directed design. *Journal for Artificial Intelligence in Engineering Design, Analysis and Manufacturing*, 9:219–230, 1995.

[7] Jon Doyle and Michael P. Wellman. Preferential semantics for goals. In *Proceedings of the Ninth National Conference on Artificial Intelligence*, pages 698–703, Anaheim, 1991.

[8] J. S. Dyer. Interactive goal programming. *Management Science*, 19:62–70, 1972.

[9] Simon French. *Decision Theory*. Halsted Press, New York, 1986.

[10] Vu Ha and Peter Haddawy. Toward case-based preference elicitation: Similarity measures on preference structures. In *Proceedings of the Fourteenth Conference on Uncertainty in Artificial Intelligence*, pages 193–201, Madison, WI, 1998.

[11] Ronald A. Howard and James E. Matheson, editors. *Readings on the Principles and Applications of Decision Analysis*. Strategic Decision Group, Menlo Park, CA, 1984.

[12] James P. Ignizio. *Linear Programming in Single and Multiple Objective Systems*. Prentice-Hall, Englewood Cliffs, 1982.

[13] R. L. Keeney and H. Raiffa. *Decisions with Multiple Objectives: Preferences and Value Trade-offs*. Wiley, New York, 1976.

[14] Yezdi Lashkari, Max Metral, and Pattie Maes. Collaborative interface agents. In *Proceedings of the Twelfth National Conference on Artificial Intelligence*, pages 444–449, Seattle, 1994.

[15] Hien Nguyen and Peter Haddawy. The decision-theoretic video advisor. In *AAAI-98 Workshop on Recommender Systems*, pages 77–80, Madison, WI, 1998.